\documentclass[letterpaper, 10 pt, conference]{ieeeconf}  %

\IEEEoverridecommandlockouts                              %

\usepackage{cite}
\usepackage{amsmath,amssymb,amsfonts}
\usepackage{algorithmic}
\usepackage{graphicx}
\usepackage{textcomp}
\usepackage{xcolor}
\usepackage{todonotes}
\usepackage{balance}
\usepackage{comment}
\usepackage[ruled,vlined]{algorithm2e}
\usepackage{subcaption}
\usepackage{tensor}
\newcommand{\eg}{\emph{e.g.},}
\newcommand{\ie}{\emph{i.e.},}

\usepackage{hyperref}
\usepackage{float}
\usepackage{multirow}
\usepackage{booktabs}

\def\secref#1{Sec.~\ref{#1}}
\def\figref#1{Fig.~\ref{#1}}
\def\tabref#1{Table~\ref{#1}}
\def\eqref#1{Eq.~(\ref{#1})}

\title{\LARGE \bf
Deep Robust Multi-Robot Re-localisation in Natural Environments 
}

\author{Milad Ramezani$^{1}$, Ethan Griffiths$^{1,2}$, Maryam Haghighat$^{2}$, Alex Pitt$^{1}$, Peyman Moghadam$^{1,2}$ 
\thanks{$^{1}$Authors are with the Robotics and Autonomous Systems, DATA61, CSIRO, Brisbane, QLD 4069, Australia. E-mails: {\tt\footnotesize \emph{
firstname.lastname}@data61.csiro.au}
$^{2}$Authors are with the Queensland University of Technology (QUT), Brisbane, Australia. E-mails: {\tt\footnotesize \emph{firstname.lastname}@qut.edu.au} }
}%

\begin{document}

\maketitle
\thispagestyle{empty}
\pagestyle{empty}

\begin{abstract}
The success of re-localisation has crucial implications for the practical deployment of robots operating within a prior map or relative to one another in real-world scenarios. Using single-modality, place recognition and localisation can be compromised in challenging environments such as forests. To address this, we propose a strategy to prevent lidar-based re-localisation failure using lidar-image cross-modality.
Our solution relies on self-supervised 2D-3D feature matching to predict alignment and misalignment.
Leveraging a deep network for lidar feature extraction and relative pose estimation between point clouds, we train a model to evaluate the estimated transformation. A model predicting the presence of misalignment is learned by analysing image-lidar similarity in the embedding space and the geometric constraints available within the region seen in both modalities in Euclidean space.
Experimental results using real datasets (offline and online modes) demonstrate the effectiveness of the proposed pipeline for robust re-localisation in unstructured, natural environments.

\end{abstract}
\section{Introduction}

(Re)-localisation in robotics refers to a process of determining a robot's current pose (position and orientation) in a known environment that has been previously mapped. This task is crucial for robots to perform their operations seamlessly, even if they experience temporary difficulty in tracking their location. 
For example, the ``wake-up'' problem involves a robot that needs to determine its location after being turned off or losing power. 
Despite significant advances in learning-based approaches for re-localisation that rely on vision~\cite{galvez2012bags, li2023hot, doan2019scalable, zhang2023etr} or lidar data~\cite{uy2018pointnetvlad, jacek20minkloc, vidanapathirana2022logg3d, hui2021pyramid},  designing a robust and reliable re-localisation technique remains a challenge, especially in unstructured, natural environments. 
Such environments lack distinctive features and change over time due to vegetation growth and weather conditions, affecting re-localisation's robustness~\cite{knights2022wild}.

Due to the inherent limitations of both lidar and image, it is difficult to extract appropriate features (in complex natural scenes) relying on a single modality for re-localisation. To address this, we propose integrating a self-supervised image-to-lidar feature matching process to predict re-localisation failure in a pipeline consisting of three modules of place recognition, pose estimation, and hypothesis verification, each leveraging learning methods. 
For place recognition and pose estimation modules, we use EgoNN~\cite{komorowski2021egonn}, an end-to-end deep re-localisation network. With the power of our lidar SLAM system, namely Wildcat~\cite{ramezani2022wildcat}, we generate lidar submaps and a pose graph with the robot nodes and geometric information in between and store them in a database. EgoNN is trained on Wildcat submaps offline. At inference time, the pre-trained network is used for re-localisation by comparing a query submap with the submaps in the database. Once the relative pose between the query and top-candidate submaps is estimated (\figref{fig:hero}), the proposed hypothesis verification module evaluates the correctness of the transformation with a cross-modality comparison between an image captured at the same time as the query submap and the top-candidate submap. 
Experimental results demonstrate the effectiveness of the proposed pipeline in achieving accurate re-localisation. The main contributions of this work can be summarised as follows:

\begin{figure}[t]
    \centering
    \subfloat{\includegraphics[width=1.0\linewidth]{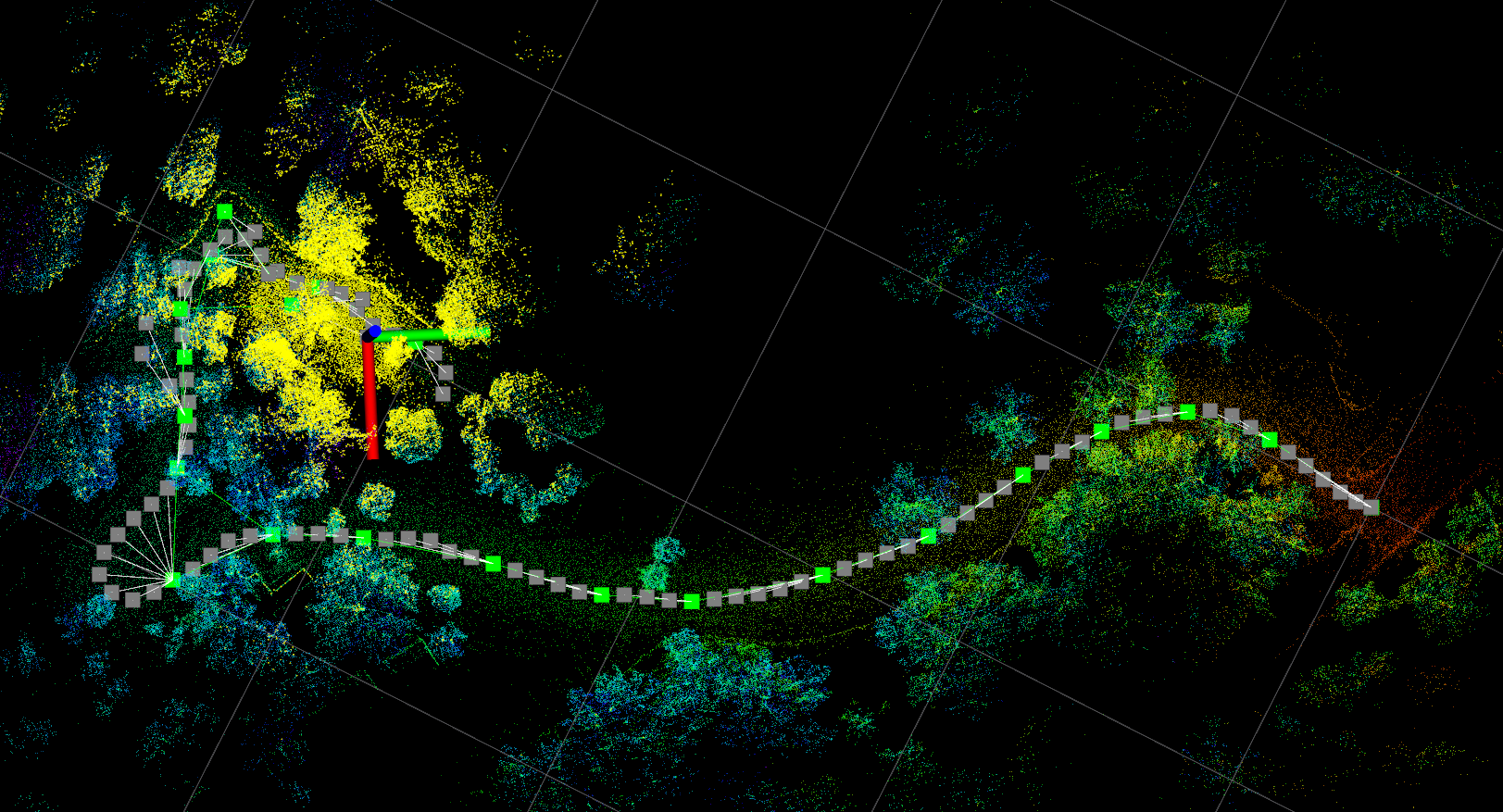}}\\
    \vspace{0.1cm}
    \subfloat{\includegraphics[width=1.0\linewidth]{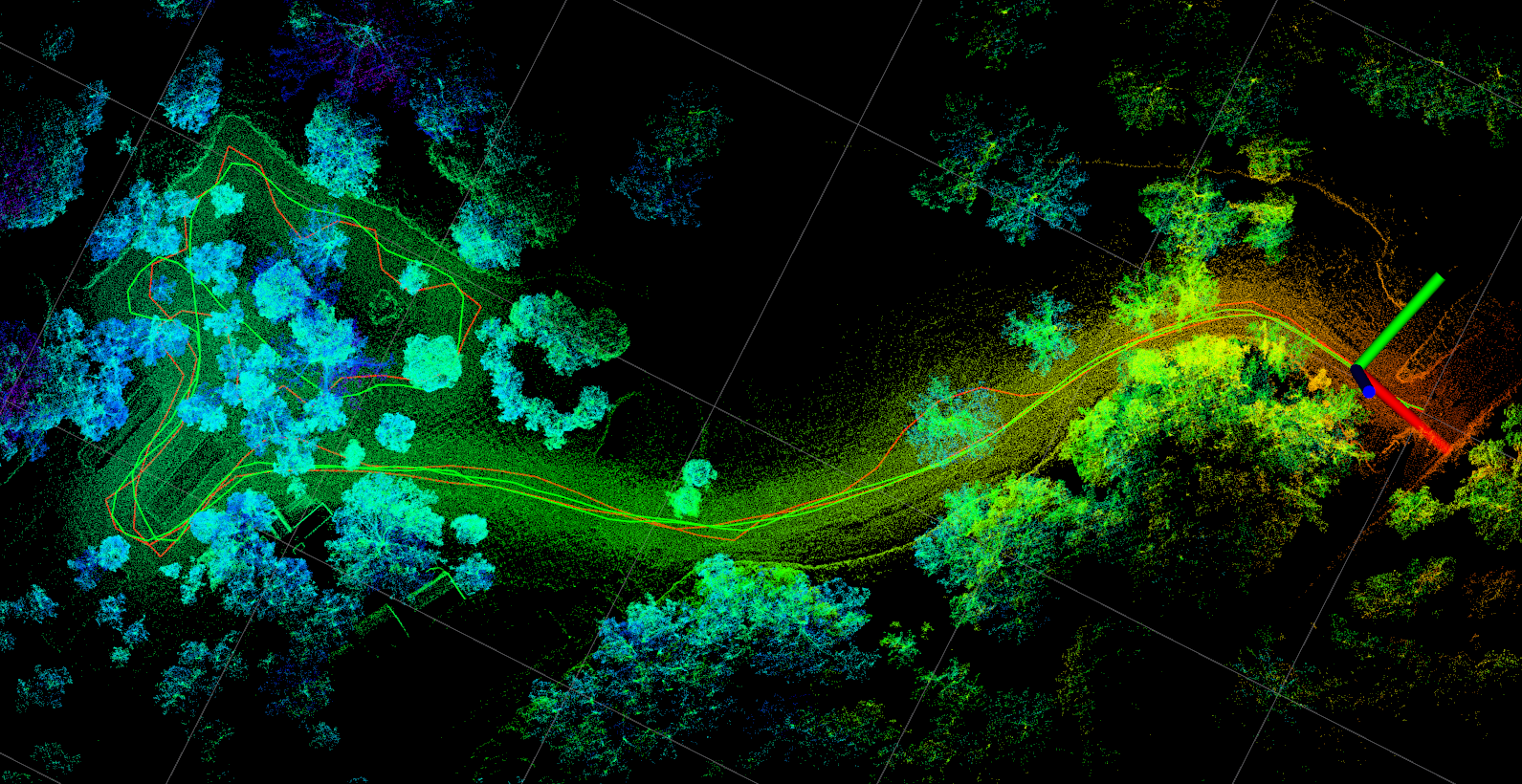}}
    \caption{\small{\textbf{Top}: The robot from the~\textit{revisit} session (location is indicated by RGB axes and query submap in yellow)} is re-localised within a pose-graph map generated by a \textit{initial} session. Re-localisation is done only based on root nodes (green) consisting of the frames of children nodes (grey). \textbf{Bottom}: Upon acceptance of the estimated relative transformation, the recent node is merged into the existing pose-graph map, allowing the \textit{revisit} session robot to continue the remaining tasks. Each grid cell is $50~\text{m}^2$.}
\label{fig:hero}
\vspace{-0.5cm}
\end{figure}

\begin{itemize}
    \item We propose integrating a self-supervised image-to-lidar feature matching process to predict re-localisation failure.
    \item We present a full pipeline of a deep re-localisation method (\textit{R3Loc}) to address multi-robot re-localisation.
    \item We demonstrate our pipeline's effectiveness in a large-scale natural dataset offline and in a forest-like environment on real robots online.
\end{itemize}

\section{Related Work}

This section reviews the existing Lidar Place Recognition (LPR) algorithms and discusses existing research into re-localisation. Finally, works related to image-lidar modal perception in cross-modal PR and registration are reviewed. 

\subsection{Lidar-Based Localisation}

A range of algorithms has been proposed for LPR. Conventional approaches~\cite{kim2018scan, rusu2009fast, rusu2008aligning, salti2014shot, dube2017segmatch} encode point clouds into either a global descriptor representing the entire point cloud or several local descriptors by segmenting the point cloud into patches. However, these handcrafted methods are often rotation dependent and are not effective in generating discriminative descriptors for unstructured environments.

Deep LPR has demonstrated outstanding results in the past few years. These methods process point clouds through a deep neural network to extract local features. Features are either directly used for place recognition, such as works in~\cite{dube2018segmap, tinchev2019learning} or aggregated using either a first-order pooling technique,~\eg~GeM~\cite{radenovic2018fine}, NetVLAD~\cite{arandjelovic2016netvlad} or second order pooling employed in~\cite{vidanapathirana2021locus, vidanapathirana2022logg3d}, to generate a global descriptor of the point cloud~\cite{vidanapathirana2022logg3d, jacek20minkloc, hui2021pyramid, uy2018pointnetvlad, zhang2019pcan}. Methods such as EgoNN~\cite{komorowski2021egonn} and LCDNet~\cite{cattaneo2022lcdnet} estimate relative pose between two point clouds upon place recognition. EgoNN computes keypoint coordinates, local descriptors, and saliency in a local head. It later estimates 6DoF relative transformation between the query and top-candidate point clouds by matching keypoints and employing RANSAC to remove outliers. LCDNet trains local features end-to-end utilising the Optimal Transport (OT) theory %
for matching features and finally estimating the relative pose using Singular Value Decomposition (SVD), allowing the entire pipeline to be differentiable, therefore, learnable. However, at test-time, LCDNet employs RANSAC for relative pose estimation prone to divergence in natural environments. 
Focusing on re-ranking top-k retrieval candidates, SpectralGV~\cite{vidanapathirana2023spectral} introduces a computationally efficient spectral re-ranking method to improve localisation.

\subsection{Cross-Modal Localisation}

There are PR-related works that aim to enhance place recognition by leveraging lidar scans and images captured in the same place. Works such as~\cite{yin2022bioslam, bernreiter2021spherical, pan2021coral} integrates lidar and visual measurements at an early stage of multi-modal fusion to encode them into a global descriptor using a projection technique; however, at the cost of dimension loss. In contrast, works such as~\cite{komorowski2021minkloc++, xie2020large, ratz2020oneshot} encode lidar and visual data separately (late fusion) into image and point cloud embeddings and later aggregate them to create the bimodal global descriptor. To deal with lighting conditions (affecting the quality of image features), AdaFusion~\cite{lai2022adafusion} employs an attention mechanism avoiding two modalities to be considered equally important when image quality is poor for recognition and vice versa.

In computer vision, works such as I2P~\cite{li2021deepi2p} and 2D3D-MatchNet~\cite{feng20192d3d} have been proposed with a focus on image-to-lidar registration. I2P trains a network to estimate the pose between a pair of images and point clouds in two steps of classification and inverse camera projection. I2P uses an attention mechanism to classify lidar points in and out of the camera frustum. It optimises pose in the lidar frame using inverse camera projection and classification prediction. 2D3D-MatchNet learns 2D image and 3D point cloud descriptors in a triplet loss (anchor image, positive and negative point clouds) as similar image-lidar descriptors are pushed closer while negative pairs are pushed apart. Recently, SLidR~\cite{sautier2022image} proposed to find similarities between point cloud and image pairs based on locally similar regions on 2D images and their corresponding 3D patches obtained knowledge distillation.  

\section{(R3Loc): Deep \underline{R}obust Multi-\underline{R}obot \underline{R}e-\underline{Loc}alisation}
\label{sec:drrr}

Our aim is to improve the robustness and reliability of (re)-localisation of a robot within a \textit{revisit} session based on a prior (reference) map generated at the \textit{initial} session by a fleet of robots in unstructured, natural environments.

Our prior map, created by Wildcat SLAM~\cite{ramezani2022wildcat}, is a pose graph $\mathcal{G}=(\mathcal{V}, \mathcal{E})$ consisting of robots' poses (nodes) $\mathcal{V}\in\mathbb{R}^6$ and the edges $\mathcal{E}\in SE(3)$ in between. In short, Wildcat integrates lidar and inertial measurements within a sliding-window localisation and mapping module. This module uses a continuous-time trajectory representation to reduce map distortion caused by motion. Undistorted sub-maps are further used in pose-graph optimisation to remove drift upon loop closure. Generated sub-maps $\mathcal{S}_i, i\in \{1, ..., n\}$ are also stored in the prior map. Further details can be found in Wildcat paper~\cite{ramezani2022wildcat} and the references therein. 

\begin{figure*}[t]
    \centering
    \includegraphics[width=1.0\linewidth]{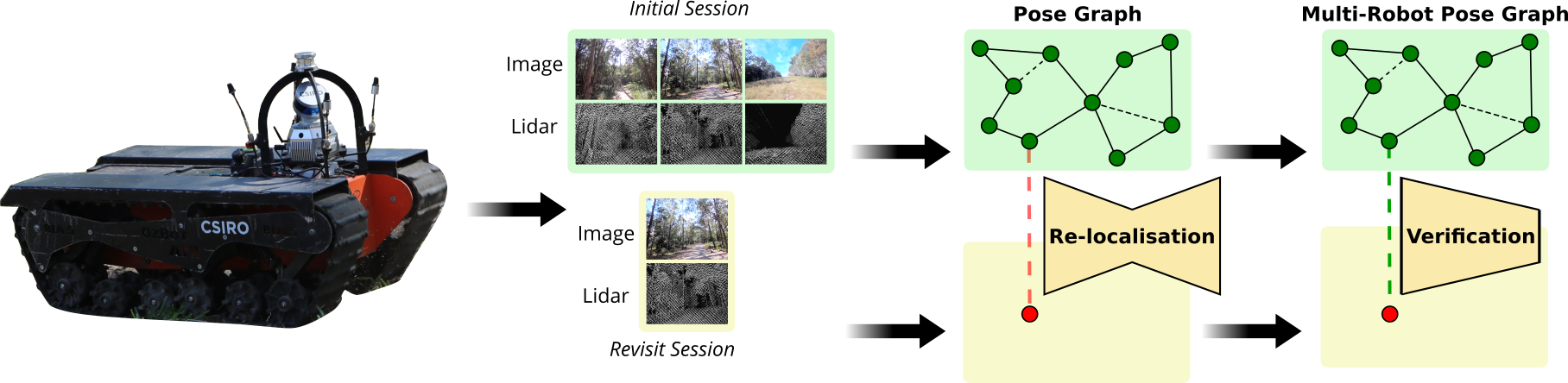}
    \caption{\small{Block diagram of the proposed deep \underline{R}obust multi-\underline{R}obot \underline{R}e-\underline{Loc}alisation system 
    (\textit{R3Loc}).}}
\label{fig:diagram}
\vspace{-0.5cm}
\end{figure*}

After generation of a new sub-map,~\ie~query point cloud $\mathcal{S}^q$, from the \textit{revisit} session, a deep lidar PR network, described in~\secref{sec:egonn}, is used to compare $\mathcal{S}^q$ with all the submaps $\mathcal{S}_i$ of the prior map to find the top candidate, $\mathcal{S}^{t1}$, using a similarity metric. Initial relative pose $\mathbf{T}_{t1,q}\in SE(3)$ between sub-maps $\mathcal{S}^q$ and $\mathcal{S}^{t1}$ is further estimated using corresponding keypoints (See~\secref{sec:egonn}) through RANSAC\cite{fischler1981random}. This initial guess is later refined with ICP, an iterative algorithm for 3D shapes registration\cite{besl1992method}. However, it needs to be evaluated before use to merge the new node into the pose graph. A false-positive edge can result in an inferior trajectory being produced or optimisation failure in SLAM.

To sanity check the refined relative pose, we propose a comparison between the query image $\mathcal{I}^q \in\mathbb{R}^{3\times W\times H}$ ($W$ and $H$ are the image width and height),~\ie~the image obtained at the same time as point cloud $\mathcal{S}^q$, and the point cloud $\mathcal{S}^{t1}$ using the estimated relative pose. To this end, we train a self-supervised network to detect 2D and 3D corresponding features and investigate the correctness of the PR output. Furthermore, we project 3D keypoints of $\mathcal{S}^{t1}$ onto the image $\mathcal{I}^q$ using the relative pose to check whether the image-lidar correspondences fall in the same region of the image. If so, the relative pose will pass to the SLAM system merging the new edge $\mathcal{E}_{t1,q}$ into the pose graph (prior map). Otherwise, we reject the relative pose.~\secref{sec:slidr} details the hypothesis verification. \figref{fig:diagram} overviews our \textit{R3Loc} pipeline, its components, and their relationship. %

\subsection{Deep Re-localisation Module}
\label{sec:egonn}
Our deep re-localisation module is based upon EgoNN~\cite{komorowski2021egonn}. Using a light 3D CNN network, EgoNN trains a global descriptor $d_{\mathcal{G}}\in \mathbb{R}^{256}$ and several local embeddings $d_{\mathcal{L}_t}\in \mathbb{R}^{128}$, where $t\in\{1,...,M\}$ is the number of keypoints detected by USIP~\cite{li2019usip}, in each point cloud.  Global descriptors are the aggregation of feature maps $\mathcal{F}_{\mathcal{G}}\in \mathbb{R}^{K\times 128}$ elements in the global head utilising GeM pooling~\cite{radenovic2018fine}. $K$ is the number of local features in the global head. Keypoint descriptors are generated in the local head by processing the elements of a local feature map $\mathcal{F}_{\mathcal{L}}\in \mathbb{R}^{M\times 64}$. A two-layer Multi-Layer Perceptron (MLP) followed by a $\tanh{}$ functional module is used to compute the local embeddings' coordinates in each point cloud. Global descriptors are used for PR, while local descriptors for localisation.

\subsection{Deep Hypothesis Verification}
\label{sec:slidr}
To accept or reject the re-localisation module output, we leverage cross-modal perception to compare the image $\mathcal{I}^q$ captured at the time of the query point cloud $\mathcal{S}^q$ and the top candidate point cloud $\mathcal{S}^{t1}$ estimated by re-localisation module. For this, the top candidate needs to be projected onto the query image using the relative pose $\mathbf{T}_{t1,q}$ estimated by local branch. If the pose estimate is correct, the projected points must overlay with their corresponding image pixels. To evaluate this, corresponding 2D and 3D features must be extracted and matched. 

Handcrafted approaches such as~\cite{sattler2016efficient} are not, however, suitable for feature extraction on lidar point clouds due to their sparsity and for the detection of similar features on images to establish accurate point-to-pixel matches. Point-wise deep feature descriptors,~\eg~\cite{feng20192d3d, li2021deepi2p}, despite outperforming conventional techniques, can be affected in the presence of occlusion or motion blur, which is inevitable in robotics.
Hence, we leverage a deep image-to-lidar self-supervised distillation approach called Superpixel-driven Lidar Representations (SLidR) \cite{sautier2022image}, which relates a group of pixels with a group of points.

SLidR trains a 3D point representation using visual features for semantic segmentation and object detection. Cross-modal representation learning is motivated by the scarcity of annotated 3D point data and the abundance of image labels. SLidR transfers feature knowledge from super-pixels,~\ie~regions of the image with visual similarity, to superpoints,~\ie~groups of points segmented through superpixels back-projection.  The image $\mathcal{I}_q$ is segmented into, at most, 250 superpixels using SLIC~\cite{achanta2012slic}. Importantly, SLidR requires no data labels for pre-training the 3D network.
Given a synchronous lidar and camera data stream and the calibration parameters, SLidR extracts features of superpixels and their corresponding superpoints. The 2D features extracted from a pre-trained ResNet-50 backbone trained using ~\cite{chen2020improved},  serve as a supervisory signal for training a 3D sparse residual U-Net backbone~\cite{choy20194d} using a contrastive loss to align the pooled 3D points and 2D pixel features.

Employing SLidR, our approach compares the extracted features of superpixels ${sp}^{\mathcal{I}_q}_i$, where $i$ is the number of superpixels in image $\mathcal{I}_q$, with that of superpoints ${sp}^{\mathcal{S}_{t1}}_j$, where $j$ is the number of superpoints in point cloud $\mathcal{S}_{t1}$, using cosine similarity:
\begin{equation}
\label{eq:cs}
    cs_{ij} = \frac{\langle \mathbf{f}_i^{\mathcal{I}_q} \; , \; \mathbf{g}_j^{\mathcal{S}_{t1}} \rangle}{\|\mathbf{f}_i^{\mathcal{I}_q} \| \| \mathbf{g}_j^{\mathcal{S}_{t1}} \|}, 
\end{equation}
here $\mathbf{f}$ and $\mathbf{g}$ denote superpixel and superpoint features, respectively, after average pooling. Symbol $\langle .,. \rangle$ denotes inner product and $\|.\|$ L2 norm. 

Now, we define two metrics, one in feature space and one in Euclidean space, to accept or reject re-localisation. First, we use the Mean Cosine Similarity (MCS) of corresponding superpixels and superpoints features,~\ie~$\frac{1}{L}\sum_{i=j}{cs_{ij}}$ to decide whether the point clouds $\mathcal{S}^{q}$ and $\mathcal{S}^{t1}$ represent the same place. $L$ is the total number of superpixel-superpoint pairs on the main diagonal of the similarity matrix computed from~\eqref{eq:cs}. Low MCS values are an indicator of false-positive cases from our re-localisation module.

Second, to evaluate the accuracy of the relative pose estimated by EgoNN, we identify the top-5 candidate superpoints, denoted as $sp^{\mathcal{S}_{t1}}$@5, for each superpixel $sp^{\mathcal{I}_q}$.
We project the centroid of each of these top-5 superpoints $sp^{\mathcal{S}_{t1}}$@5 onto the image $\mathcal{I}_q$.
We find the superpoint $sp_c^{\mathcal{S}_{t1}}$ whose projected centroid is closest to the centroid of $sp^{\mathcal{I}_q}$, and we select it as the pair of $sp^{\mathcal{I}_q}$. 
We check whether the projected centroid of $sp_c^{\mathcal{S}_{t1}}$ falls within $sp^{\mathcal{I}_q}$, and we count it as a match if it does and a mismatch if it does not.
We calculate the percentage of superpixel-superpoint mismatched pairs over the entire set of pairs to determine whether to reject or accept the re-localisation. We then define the \emph{alignment} ratio as follows:
\begin{equation}
    \label{eq:confirmity}
    \nu = 1 - \frac{n}{L},
\end{equation}

where $n$ is the number of superpixel-superpoint mismatched pairs computed from the abovementioned procedure. 
Defining two similarity and alignment metrics, we train a simple multi-class Support Vector Classifier (SVC),  $y_i=\mathcal{K}(\text{MCS}_i, \nu_i)$, to predict if the pair $i$ belongs to matched, mismatched or unmatched category, where $y_i\in\{\text{matched, mismatched, unmatched}\}$.

\section{Experimental Results}
In this section, we present the following results: evaluation of the re-localisation module on a large-scale natural dataset, Wild-Places~\cite{knights2022wild} (consisting of \emph{Venman} and \emph{Karawatha} sequences) and its comparison with Scan Context~\cite{kim2018scan} (as a handcrafted PR approach widely integrated with lidar SLAM), evaluation of cross-modal localisation on the same dataset. Finally, we evaluate the entire proposed \textit{R3Loc} pipeline on a wake-up problem scenario on a robot system. 

Both EgoNN and SLidR were trained on the Wild-Places dataset. For EgoNN we followed the training split described in~\cite{knights2022wild}. For testing, however, we evaluated the model on two sequences of Venman collected in opposite directions. This inter-sequence PR evaluation mimics the wake-up problem when the robot operating in the \textit{revisit} session travels within the prior map generated in the opposite direction. Same sequences were used to evaluate Scan Context following the default setting. For evaluation, we define a true positive revisit when a prediction is within 3 m of a positive ground truth.

For SLidR we trained and validated the network using about 1750 matched lidar-image pairs (pairs captured at the same time)
on one sequence from Venman. 
We created three test sets from the validation section by augmenting the relative transformation between the image and point cloud to create matched and mismatched pairs, and by randomly pairing images and point clouds captured in different places for unmatched pairs. This allows testing SLidR for the three most common EgoNN output cases. We also tested the proposed verification pipeline on a new dataset collected in an unstructured area at the Queensland Centre for Advanced Technology (QCAT), Brisbane, Australia.

\subsection{Evaluation of EgoNN Offline}
\begin{figure}[b]
    \centering
    \includegraphics[width=0.8\linewidth]
    {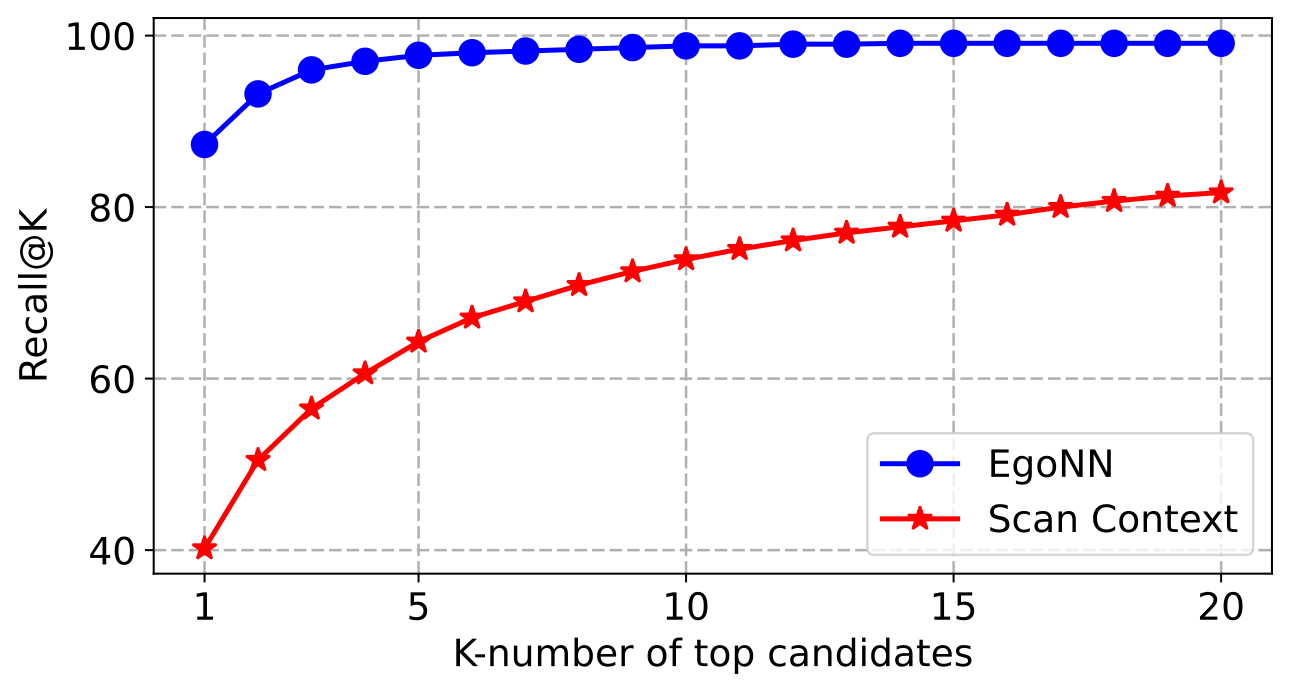}
    \caption{\small{Recall@K performance between EgoNN and Scan Context.}}
\label{fig:recall}
\end{figure}

\figref{fig:recall} illustrates the top-K Recall curves between EgoNN and Scan Context. As seen, the performance of EgoNN is almost twice higher than Scan Context, indicating the limitation of Scan Context to produce distinctive and rotation-invariant descriptors in cluttered environments such as forests. To evaluate re-localisation accuracy, we compare the estimated relative transformation with ground truth and compute a success rate if the rotation and translation errors are within $5^{\circ}$ and $2$~m, respectively. This evaluation was not performed for Scan Context due to the inability of this approach to estimate 6DoF rotation and translation only based on global descriptors. 

The success rate for EgoNN, when the relative transformation was only estimated using keypoints and via RANSAC, is about $40\%$. However, after refining the estimated transformation using ICP (we downsampled point clouds to 40 cm spatial resolution for online registration), the success rate increased to $78\%$. This indicates that although EgoNN achieved high performance in place recognition, the extracted keypoints are not well-repeated in unstructured environments for accurate re-localisation.

\begin{figure*}[t]
    \centering
    \includegraphics[width=0.9\linewidth]
    {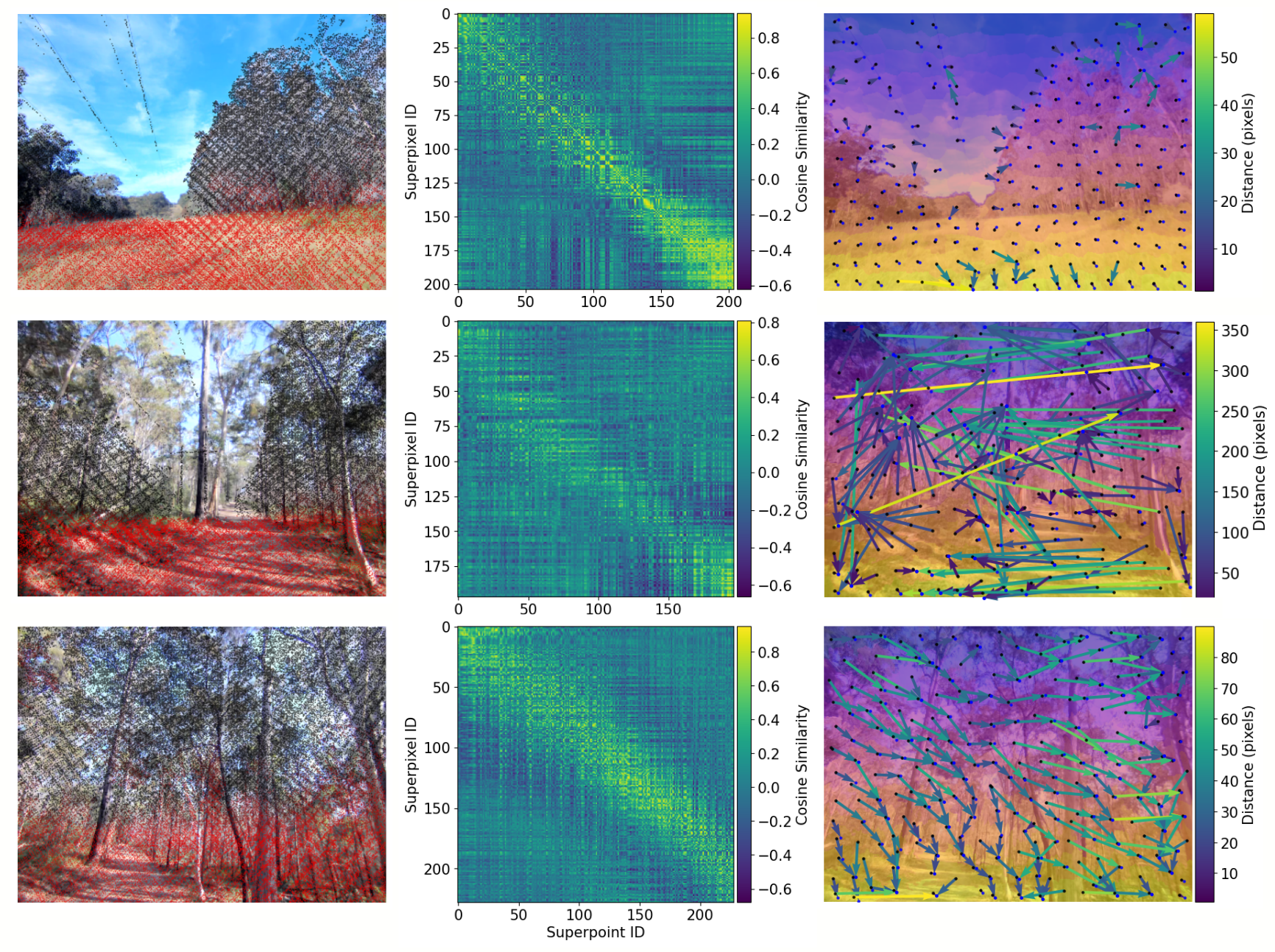}
    \caption{\small{Examples of matched (top), unmatched (middle) and mismatched (bottom) pairs. From left to right: projected points on the image, superpoint-superpixel similarity matrix, projected error vectors based on top candidate in the similarity matrix.  When the image and point cloud are captured in the same place, our proposed verification metrics, \ie{} Mean Cosine Similarity (MCS) and alignment ratio ($\nu$) (calculated through the process described in~\secref{sec:slidr}) are used to identify true/false-positive PR and predict the re-localisation success or failure correspondingly. 
    }}
\label{fig:slidr}
\vspace{-0.5cm}
\end{figure*}

\begin{figure}[t]
    \centering
    \includegraphics[width=1.0\linewidth]{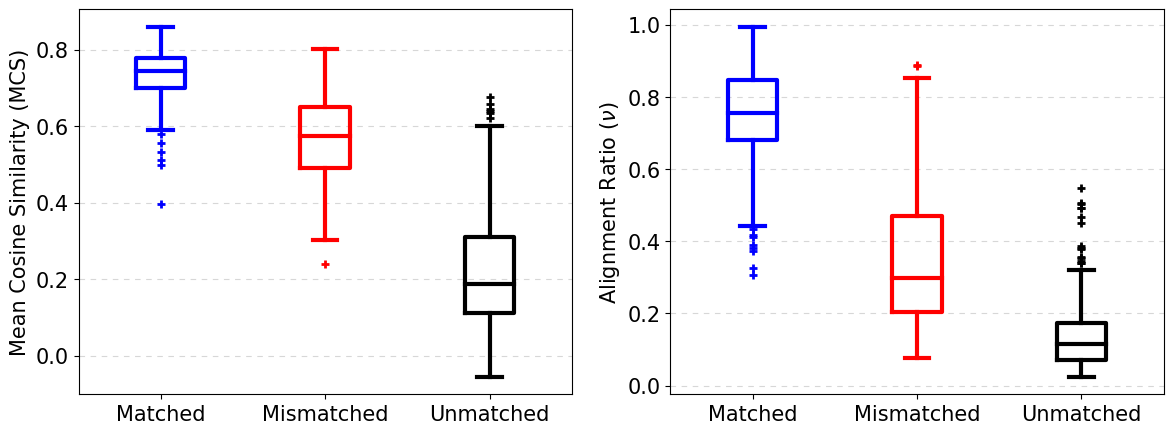}
    \caption{\small{Boxplots of statistical analysis of MCS and $\nu$ on the matched, mismatched and unmatched pairs in the validation set.}}
\label{fig:valid}
\vspace{-0.5cm}
\end{figure}

\subsection{Evaluation of SLidR Offline}
\figref{fig:slidr} illustrates an example of matched, unmatched and mismatched pairs in the top, middle and bottom rows, respectively. As seen, the similarity matrices (second column) and the projection vectors (third column) obtained from the procedure described in~\secref{sec:slidr} are good measures to distinguish between matched, mismatched and unmatched pairs.~\figref{fig:valid} shows the boxplots over MCS and $\nu$ computed for about 250 matched pairs and 230 mismatched and unmatched pairs on the validation set (above 700 pairs in total). The substantial difference in MCS between unmatched and matched/mismatched pairs allows classifying unmatched ones with high confidence. Additionally, the large $\nu$ for the matched pairs helps classify them from the other pairs. We, however, observed if both MSC and $\nu$ are used together, it improves generalisation when training and testing environments are different. Hence, we trained  a multi-class fifth-degree polynomial SVC model, $\mathcal{K}(\text{MCS}, \nu)$, to predict if a pair belongs to matched, mismatched or unmatched categories.

\begin{figure}[t]
    \centering
    \includegraphics[width=0.9\linewidth]{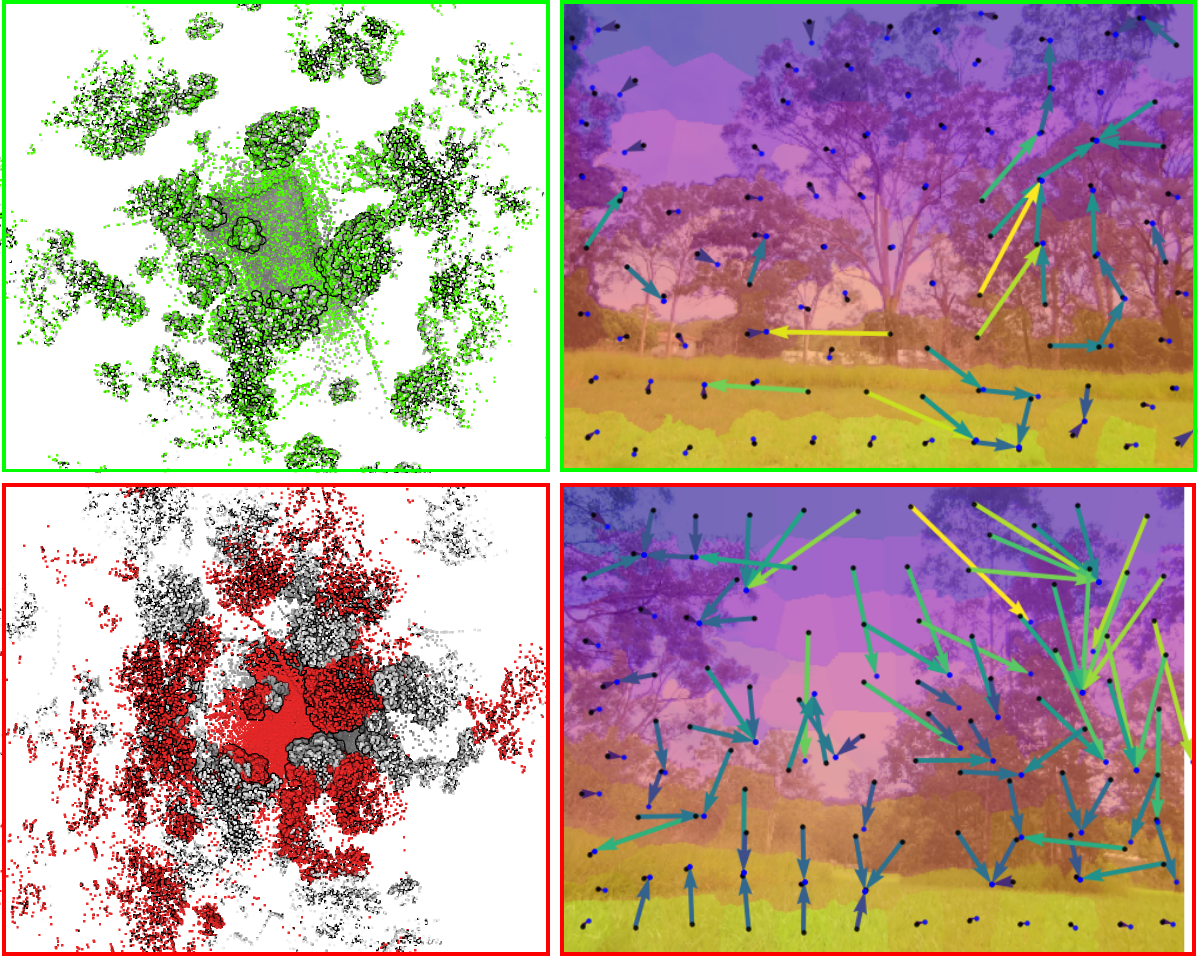}
    \caption{\small{Re-localisation success (top) and failure (bottom) cases in the QCAT dataset.}}
\label{fig:qcat_examples}
\vspace{-0.7cm}
\end{figure}

\subsection{Evaluation of the Entire Pipeline Online}
To evaluate our pipeline, a tracked robot (as seen in~\figref{fig:diagram}), equipped with a lidar sensor and four cameras, was teleoperated in an unstructured area once as the \textit{initial} session and once as the \textit{revisit} session at QCAT. Time difference between the two sessions was reasonably chosen large, allowing us to evaluate the performance of our verification under various lighting conditions.
For the cross-modal perception, we only used the camera frames of the front camera. Submaps were generated by Wildcat, and Robot Operating System (ROS) was used for communication between different components. Our re-localisation pipeline is triggered through a rosservice command. Upon requesting re-localisation, the query submap and the submaps existing in the prior map were fed into the already trained model of EgoNN. By performing a forward pass using the weights and benefiting from kd-tree, the top candidate was selected and the relative pose was estimated. Since PR is only performed based on root nodes, there were at most 20 submaps in the prior map from \textit{initial} session. To thoroughly test the pipeline, the recorded data of the \textit{revisit} session played back, and the re-localisation was requested for every root node spawned out, resulting in testing the entire pipeline 20 times (\ie~20 ``wake-up'' locations). Following this process, the average Recall@1 of EgoNN is $100\%$. However, the success rate for re-localisation was $70\%$, evidencing the necessity of the hypothesis verification. 

The pose estimate is not transferred to our lidar-inertial SLAM unless it passes through the hypothesis verification. For this, the top-candidate submap and the query image (which is already rectified) are fed into our pre-trained verification models. For the QCAT dataset, after 20 trials, the proposed hypothesis verification detected all the mismatched pairs for which EgoNN could not produce an accurate pose estimate.~\figref{fig:qcat_examples} shows a sample matched (top) and mismatched (bottom) scenario. The proposed verification pipeline, including the pre-trained feature matching  and the SVC model $\mathcal{K}$,  successfully separated these cases and detected the re-localisation failure.

\begin{figure*}[t]
    \centering
    \includegraphics[width=0.95\linewidth,trim={0cm 0.2cm 0 0cm},clip]{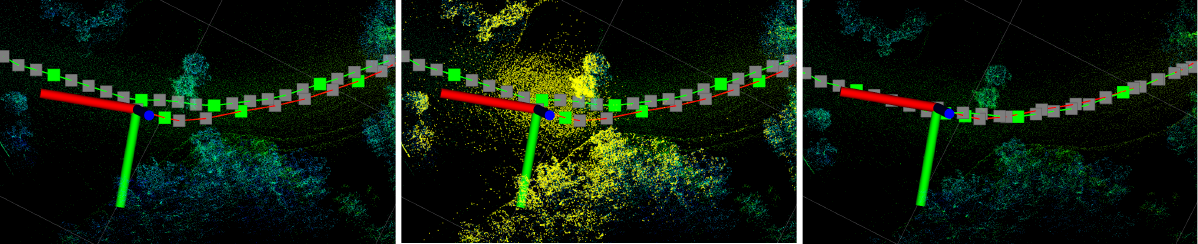}
    \caption{\small{Re-localisation example. When a robot (from the \emph{revisit} session) moves in a known environment (the map generated from the \emph{initial} session) (left), upon a successful re-localisation (middle), the current pose graph is merged into the prior map (right), allowing the robot to operate concerning the prior map for resuming unfinished tasks.}}
\label{fig:qcat_multiagent}
\vspace{-0.2cm}
\end{figure*}

\begin{figure}[t]
    \centering
    \includegraphics[width=0.9\linewidth,trim={0cm 0.4cm 0 0cm},clip]{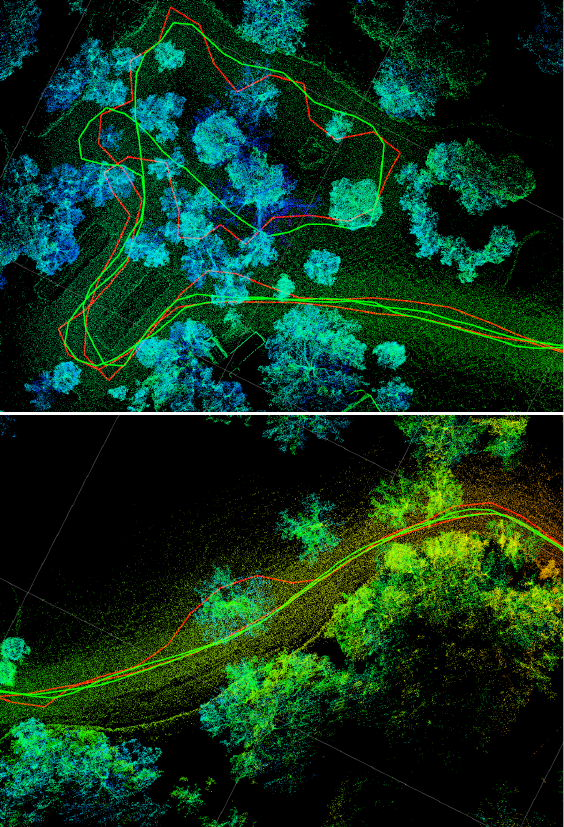}
    \caption{\small{Multi-agent re-localisation. Our system can be used to merge the pose graph created by individual agents (paths shown in red and green) operating simultaneously in an environment.}}
\label{fig:qcat_multiagent_paths}
\vspace{-0.5cm}
\end{figure}

Upon a verified re-localisation, the pose graph generated by the \emph{revisit} session is safely merged into the existing map as shown in~\figref{fig:qcat_multiagent}.
\figref{fig:qcat_multiagent_paths} presents qualitative results after merging \textit{revisit} session robot into the map generated from the \textit{initial} session, evidencing the proposed pipeline feasibility for multi-agent re-localisation. 

\subsection{Runtime Analysis}
\label{tab:runtime}
\begin{table}[t]
    \caption{\small{Runtime Analysis.}}
    \label{tab:runtime}
    \resizebox{\linewidth}{!}{\begin{tabular}{c|cc|cccc|c}
        \hline
        {\textbf{Processing}}& \multicolumn{2}{|c|}{EgoNN}& \multicolumn{4}{|c|}{SLidR}& {\textbf{Total}}\\ 
        \cline{2-4} \cline{5-7} 
        \multirow{1}{*}{\textbf{time}}& Description& Localisation& Superpixel & Description& MCS & Verification& \textbf{time}\\\hline
        & &  & & & && \\
         \textbf{Mean}(s)& 0.087&  0.408& 0.186 &0.209& 0.013 & 0.019 & 0.905\\
         \textbf{std}(s) & $\pm$0.002& $\pm$0.207& $\pm$0.085 & $\pm$0.022 &$\pm$0.037 & $\pm$0.002 & $\pm$0.215\\
        \hline
    \end{tabular}}
    \vspace{-0.5cm}
\end{table}

To demonstrate that our presented system can run online, we evaluated the computation time for each
component. The timing results are collected by running the pre-trained models on a single NVIDIA Quadro T2000 GPU and the rest of the pipeline on a unit with an Intel Xeon W-10885M CPU.
~\tabref{tab:runtime} reports a breakdown of individual modules' runtime in our pipeline. 
Together, the total runtime (for the QCAT experiment with the scale shown in~\figref{fig:hero}) is less than a second, allowing the system to run for online operation.

\section{Conclusion}
\label{sec:conclusion}
This work introduced a robust multi-robot re-localisation system. Our re-localisation pipeline benefits from deep lidar representation in place recognition and pose estimation. Self-supervised image-to-lidar
knowledge distillation was used to reason about the alignment between the image captured at the same time of query point cloud and the top-candidate point cloud. The system's modules were separately tested on a large-scale public dataset, and the entire pipeline, integrated with our lidar SLAM system, has been tested online in a wake-up case scenario. In future, we will further investigate how to improve the cross-modal perception by end-to-end learning of representation, including image segmentation for superpixel creation in unstructured environments, and verification models.

\section*{Acknowledgments}
The authors gratefully acknowledge funding of the project by the CSIRO's Reimagine Farming initiative. They also thank CSIRO Robotics and Autonomous Systems members for the hardware and software support, particularly Mark Cox for his insights on computer vision. M.H. acknowledges ongoing support from the QUT SAIVT research program.
\balance{} 
\bibliographystyle{IEEEtran}
\bibliography{refs}

\end{document}